\documentclass[runningheads]{llncs}

\usepackage{algorithm,algpseudocode}

\usepackage{setspace}
\usepackage{longtable,booktabs}
\usepackage{graphicx,grffile}
\usepackage{amssymb,amsmath}
\usepackage{url}
\usepackage{bm}

\usepackage[misc]{ifsym}

\usepackage{color}
\definecolor{BrightRoyalPurple}{rgb}{0.65, 0.05, 0.78}

\begin{document}

\title{Performing Arithmetic Using a Neural Network Trained on Digit Permutation Pairs}
\titlerunning{Performing Arithmetic Using a Model Trained on Digit Permutation Pairs}
%
\author{Marcus D. Bloice\inst{1}\textsuperscript{,(\Letter)} \and
Peter M. Roth\inst{2}\and
Andreas Holzinger\inst{1}}
\authorrunning{M.D. Bloice et al.}
%
\institute{Institute for Medical Informatics, Statistics, and Documentation\\Medical University Graz, Austria\\ \email{marcus.bloice@medunigraz.at}\\ \email{andreas.holzinger@medunigraz.at}\and
Institute of Computer Graphics and Vision\\Graz University of Technology, Austria\\ \email{p.m.roth@ieee.org}}

\maketitle

\begin{abstract}
In this paper a neural network is trained to perform simple arithmetic using images of concatenated handwritten digit pairs. A convolutional neural network was trained with images consisting of two side-by-side handwritten digits, where the image's label is the summation of the two digits contained in the combined image. Crucially, the network was tested on permutation pairs that were not present during training in an effort to see if the network could learn the task of addition, as opposed to simply mapping images to labels. A dataset was generated for all possible permutation pairs of length 2 for the digits 0--9 using MNIST as a basis for the images, with one thousand samples generated for each permutation pair. For testing the network, samples generated from previously unseen permutation pairs were fed into the trained network, and its predictions measured. Results were encouraging, with the network achieving an accuracy of over 90\% on some permutation train/test splits. This suggests that the network learned at first digit recognition, and subsequently the further task of addition based on the two recognised digits. As far as the authors are aware, no previous work has concentrated on learning a mathematical operation in this way.
\end{abstract}

\section{Introduction}
The aim of this study is to attempt to find experimental evidence that would suggest that a network can be trained to perform the task of addition, when supplied with image data containing two digits that should be summed. To ensure that the network has indeed learned this, and is not simply mapping images to labels, a constraint was applied whereby the network is tested with a held back test set of previously unseen permutation pairs. This forces the network to learn more than simply a mapping between individual images and labels as it is tested using digit combination pairs that it has not seen, meaning a direct mapping from an image or shape to a label would not function.

\begin{table}[h!]
\caption{Example input images and their corresponding labels. Each image consists of two side-by-side MNIST digits merged into one single image where each image's label is the summation of the two digits. The label contains no information as to the individual digits contained within the image, nor is there any indication given to the network prior to training that each image consists of two digits or otherwise.}
\label{table:examples}
\centering
\begin{tabular}{@{}lcc@{}}
\toprule
Image                                                                                      & Interpretation & Label \\ \midrule
\begin{tabular}[c]{@{}l@{}} \includegraphics[scale=0.5]{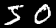} \end{tabular} & 5 + 0          & 5     \\
\begin{tabular}[c]{@{}l@{}} \includegraphics[scale=0.5]{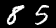} \end{tabular} & 8 + 5          & 13    \\
\begin{tabular}[c]{@{}l@{}} \includegraphics[scale=0.5]{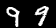} \end{tabular} & 9 + 9          & 18    \\ \bottomrule
\end{tabular}

\end{table}

To test this hypothesis, a network was trained with data generated using 90 of the possible 100 combinations of the digits 0--9 up to length 2. Once trained, the network was tested by inputting images based on the remaining  10 permutation pairs, and was required to predict the summations for them. Some examples of the generated samples can be seen in Table~\ref{table:examples}. 1,000 samples were generated for each permutation pair. This was repeated 10 times for 10 different permutation pair train/test splits.

\section{Background Work}

There has been previous work relating to this experiment. As opposed to most work, however, the goal of this study was not to recognise digits, extract their numerical values from the images, and then perform (after the network's character recognition procedure) some mathematical function on the numerical values. The task of this experiment was to learn if the network could learn the logical task of the mathematical operation itself using an end-to-end approach. For example, \cite{hoshen2016} experimented with computer generated image data, however as output the network was trained to produce images containing the summations. Their work concentrated on the visual learning of arithmetic operations from images of numbers. In contrast, the work presented here outputs its predictions as a real number. Their approach used numbers of longer lengths and were therefore also able to generate many thousands of training samples, despite not using hand written digits. The input consisted of two images, each showing a 7-digit number and the output, also an image, displayed a number showing the result of an arithmetic operation (e.g., addition or subtraction) on the two input numbers. The concepts of a number, or of an operator, are not explicitly introduced. Their work, however, was more akin to the learning of a transformation function, rather than the task of learning a mathematical operation. Other operations, such as multiplication, were not learnable using this architecture. Some tasks were not learnable in an end-to-end manner, for example the addition of Roman numerals, but were learnable once broken into separate sub-tasks: first perceptual character recognition and then the cognitive arithmetic sub-task.

Similarly, a convolutional neural network was used by \cite{liang2016} to recognise arithmetic operators, and to segment images into digits and operators before performing the calculations on the recognised digits. This again is different to the approach described here, as it is not an attempt to learn the operation itself, but to learn to recognise the operator symbols and equations (and perform the mathematics on the recognised symbols).

In \cite{WalachWolf:2016:counting}, the authors addressed the task of object counting in images where they applied a learning approach in which a density map was estimated directly from the input image. They employed convolutional neural networks with layered boosting and selective sampling. It would be possible to create an experiment based on their work, that would perform arithmetic by counting the values of domino tiles, for example.

Until now, as far as the authors are aware, no work has concentrated on learning the actual mathematical operation itself. Previous work tends to concentrate on first recognising digits and operators within images, and then to perform the mathematical operations separately, after this extraction has been carried out. In the case of this work, an end-to-end algorithm has been developed that performs the digit recognition, representation learning of the values of the digits, and performs the arithmetic operation.

\section{Experiment}
A convolutional neural network was trained to perform arithmetic on images consisting of two side-by-side hand-written digits. Each image's corresponding label is the sum of the two digits, and the network was trained as a regression problem. For the digits 0--9, up to length 2, there are 100 possible permutation pairs. For each permutation pair, 1,000 unique images were generated using the MNIST hand written digit database \cite{lecun1998}. Training was performed on images generated from a random 90-long subset of the possible permutations, and testing was performed on images based on the remaining 10 permutations. A number of example input images and their labels are shown in Table~\ref{table:examples} where it can be seen that a single input image consists of two MNIST digits side-by-side, and the image's label is the summation of the two digits. For each permutation, 1,000 combined images are generated resulting in 100,000 samples that are separated into a training and test set based on the permutation pairs.

The task of the experiment was to train a neural network with data generated from a subset of the possible 100 permutations, that when presented with a new samples, generated from the unseen permutation pairs, the network would be required to predict the correct summation. This was done in order to ascertain whether a network could learn a simple arithmetic operation such as addition, given only samples of images and their summations and no indication as to the value of each individual digit contained within the image, while only being trained on a subset of all possible permutation pairs and tested on the remaining pairs.

In summary, the experimental setting is as follows:

\begin{itemize}
  \item By permutations, it is meant all possible combinations of the digits 0--9 of length 2. Formally, if the set of digits $D = \{0,1,2,3,4,5,6,7,8,9\}$, all possible permutations is the Cartesian product of $D \times D$, which we define as $P$, so that $P=\{(0,0), (0,1), (0,2), \ldots, (9, 8), (9,9)\}$.
  \item Of the 100 possible permutations pairs $P$, a random 90 are used as a basis to train the network and the remaining 10 pairs are used as a basis to test the network. These are the training permutations, $P_t$, and the test permutations, $P_v$. This permutation train/test split was repeated 10 times as a 10-fold cross validation.
  \item For each permutation, 1,000 samples are generated. So, for each of the permutations in $P$, 1,000 concatenated images are generated using random MNIST digits $\bm{M}$ (appropriate for that permutation). 
  \item By appropriate this means that, for example, generating an image for the permutation pair $(3, 1)$ a random MNIST digit labelled 3 is chosen and a random MNIST digit labelled 1 is chosen and these images are concatenated to create a single sample for this permutation pair. This means each sample is likely unique (likely, as each image is chosen at random with replacement, see Table~\ref{table:randomness}). The generated images are contained in a matrix $\bm{X}$, where ${\bm X}_t$ are the training samples and ${\bm X}_v$ are the test samples. 
  \item For the generation of the training set images, ${\bm X}_t$, only images from the MNIST training set, ${\bm M}_t$, are used.
  \item For the test permutation images, ${\bm X}_v$, only images from the MNIST test set, ${\bm M}_v$, are used.
  \item The network is not given any label information regarding each individual digit within the concatenated images, only the summation is given as label data.
  \item The permutations pairs in the test set are not seen during training. This means the training set and test set are distinct in two ways: they contain different permutations pairs that do not overlap, and the individual MNIST images used to generate each permutation sample do not overlap between the training set and test set.
\end{itemize}

\begin{table}
\caption{For each permutation pair, random MNIST digits are used for generating each sample. For example, for the permutation pair (0, 2), each sample that is generated uses a random digit 0 combined with a random digit 2 obtained from MNIST.}
\label{table:randomness}
\centering
\begin{tabular}[]{@{}llll@{}}
\toprule
Sample 1 & Sample 2 & \ldots & Sample 1000 \tabularnewline
\midrule
\begin{tabular}[c]{@{}l@{}} \includegraphics[scale=0.5]{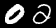} \end{tabular} &
\begin{tabular}[c]{@{}l@{}} \includegraphics[scale=0.5]{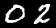} \end{tabular} &
\ldots &
\begin{tabular}[c]{@{}l@{}} \includegraphics[scale=0.5]{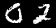} \end{tabular} \tabularnewline
\bottomrule
\end{tabular}
\end{table}

The decision to train the network as a regression problem is done for the following reasons. First, the number of output neurons would change depending on the train/test split. For example, the sum of $(9, 9)$ is 18 and cannot be made by any other permutation, meaning a possible discrepancy between the number of possible output neurons of the training set and test set. Second, when training on permutations of longer length, an ever increasing number of output neurons would be required. Last, for measuring how well the network performs, classification poses problems which are mitigated by using regression and mean squared error loss.

The following sections describe the experiment itself, beginning with a description of the dataset and how it was created. Then, the neural network's architecture is described as well as the training strategy. Last, the results of the experiment are discussed, and the paper is concluded with a discussion on future work.

\section{Dataset}
The dataset used for the creation of the concatenated image data was MNIST, a 70,000 strong collection of labelled hand written digits. As per the original dataset, 60,000 digits belong to the training set and 10,000 belong to the test set. Images in the MNIST dataset are 8-bit greyscale, $28\times28$ pixels in size. The generated images are therefore $28\times56$ pixels in size as they are the concatenation of two MNIST digits placed side-by-side and stored as a single image (examples of which are shown throughout this paper). Each generated image's label is the summation of the two individual digits' labels (see Table \ref{table:examples} for several examples). For each permutation, one thousand samples are generated, and the MNIST digits are chosen at random in order to create distinct samples.

\subsection{Train/Test Split}
As mentioned previously, for the digits 0--9, with a maximum of length of $l=2$, there are $n^l$ or $10^2=100$ possible permutation pairs. To generate the training and testing data, the permutations pairs are split into a permutation training set and a permutation test set at random, so that: $P = P_{t} \cup P_{v}$ and $P_t \cap P_v = \varnothing$. For training, 90\% of the permutations were used to generate the training samples, ${\bm X}_t$ and the remaining 10\% were used for generating the the test set samples ${\bm X}_v$, meaning $|P_{t}|=90$ and $|P_{v}|=10$ while $|{\bm X}_{t}|= 90,000$ and $|{\bm X}_{v}|= 10,000$ for any particular run. The experiment was performed using a 10-fold cross validation, based on the permutations pairs, and the loss was averaged across the 10 runs.

In terms of the generated samples, the generated training set images and generated test set images honoured the MNIST training set and test set split. This means that the generated training set samples are distinct from the generated test set samples both in terms of the permutation pairs and the images used to create each sample. It should be noted that the images were chosen from MNIST randomly \textit{with replacement}, meaning images could appear twice in different permutation pairs within the training set or test set, but not between both. Therefore, the generated data set matrix $\bm{X}$ contains 100,000 samples, 10,000 for each permutation pair. This also means that ${\bm X}_t$ contains the corresponding samples for the permutations $P_t$, and ${\bm X}_v$ contains the samples for $P_v$. A label vector $\bm{y}$ contains the labels, which are the summations of the two digits in the sample. Similarly, ${\bm y} = {\bm y}_t \cup {\bm y}_v$.

\subsection{Network Architecture and Training Strategy}
The neural network used was a multilayer convolutional neural network similar to the original LeNet5, which was first reported by \cite{LeCunEtAl:1989:Handwritten} (see also \cite{HubelWiesel:1962:CatVisual}). However, rather than treating the problem as a classification problem, the network is trained as a regression problem. The network was evaluated using mean squared error loss and optimised with ADADELTA \cite{zeiler2012}. All experimentation was performed with Keras using TensorFlow \cite{tensorflow2016} as its back-end, running under Ubuntu Linux 14.04. Training was undertaken using an NVIDIA Titan X GPU.

A LeNet5-type network was chosen due to its association with MNIST, having been optimised and developed for this dataset, and it is an algorithm that has repeatedly been shown to work well with the general task of character recognition. As the inputs to this network are similar to the original MNIST images, having twice the width in pixels but having the same height in pixels, the only other modification that was made was with the output of the network. Instead of a 10 neuron, fully connected output layer with \textit{Softmax}, the final layer was replaced with a single output neuron and trained as a regression problem, optimising mean squared error loss.

\subsection{Data Generation}
The data generation procedure algorithm is shown in Algorithm~\ref{alg:datageneration}. During data generation, the permutations, $P$, are iterated over and $m=1000$ samples are generated for each permutation. A sample consists of two random MNIST images, corresponding to the labels in the current permutation, concatenated together as one image (this is represented by the function \texttt{ConcatenateImages}). As well as this, the label vector $\bm{y}$ is generated, which contains the sum of the two digit labels that make up each individual MNIST image used to create the sample. Note that in Algorithm~\ref{alg:datageneration} the symbol $\Leftarrow$ represents appending to the data structure, in this case the matrix $\bm{X}$ and its corresponding label vector $\bm{y}$.

The procedure shown in Algorithm~\ref{alg:datageneration} is repeated for the train set permutation pairs, $P_t$, and the test set permutation pairs, $P_v$. It is important to note that when generating the data for the training set permutation images, the MNIST training set is used, and conversely when generating the test set permutation images, the MNIST test set is used. This ensures no overlap between the training set or test set in terms of the permutation pairs or the data used to generate the samples. 

\begin{algorithm}[tb]
   \caption{Data Generation}
   \label{alg:datageneration}
\begin{algorithmic}
   \State {\bfseries Input}: permutation pairs $P$, MNIST images ${\bm M}$, labels ${\bm y'}$, size $m \leftarrow 1000$.
   \State Initialise ${\bm X} \leftarrow [\,]$.
   \State Initialise ${\bm y} \leftarrow [\,]$.
   \ForAll {$(p_1, p_2)$ {\bfseries in} $P$}
   \For{$1$ {\bfseries to} $m$}
   \State $r^1 \leftarrow$ random index from ${\bm M}$ with label $p_1$
   \State $r^2 \leftarrow$ random index from ${\bm M}$ with label $p_2$
   \State ${\bm X} \Leftarrow$ ConcatenateImages(${\bm M}_{r^1}$, ${\bm M}_{r^2}$)
   \State ${\bm y} \Leftarrow {\bm y'}_{r^1} + {\bm y'}_{r^2}$
   \EndFor
   \EndFor
\end{algorithmic}
\end{algorithm}

\section{Results}
\label{sec:results}
Averaged across the different training/test splits of a 10-fold cross validation of the permutation pairs, mean squared error was generally under 1.0 and averaged 0.85332, as shown in Table~\ref{table:crossValResults}. Tables~\ref{table:exampleResults}, \ref{table:exampleResultsSameLabel}, and \ref{table:exampleResultsSameLabel2} show a number of examples of a trained network's predictions on permutations from a test set. Table \ref{table:exampleResults} shows a number of sample inputs from the test set and their predictions, as well as their true labels. It is interesting to note that the network learned to deal with permutations with images in reverse order, as is the case for (6, 4) and (4, 6) or (1, 3) and (3, 1) in Table~\ref{table:exampleResultsSameLabel}. In some cases, three different permutation pairs exist in the test set which sum to the same number, and these were also predicted correctly, as seen in Table~\ref{table:exampleResultsSameLabel2}.

\begin{table}
\caption{Example results for permutation samples from the test set passed through the trained network. As can be seen, all samples use distinct MNIST digits.}
\label{table:exampleResults}
\centering
\begin{tabular}[]{@{}lcc@{}}
\toprule
Input Image & Prediction & Actual\tabularnewline
\midrule
\begin{tabular}[c]{@{}l@{}} \includegraphics[scale=0.5]{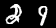} \end{tabular} & 11.1652 & 11\tabularnewline
\begin{tabular}[c]{@{}l@{}} \includegraphics[scale=0.5]{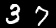} \end{tabular} & 10.3215 & 10\tabularnewline
\begin{tabular}[c]{@{}l@{}} \includegraphics[scale=0.5]{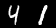} \end{tabular} & 5.01775 & 5\tabularnewline
\begin{tabular}[c]{@{}l@{}} \includegraphics[scale=0.5]{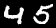} \end{tabular} & 8.99357 & 9\tabularnewline
\begin{tabular}[c]{@{}l@{}} \includegraphics[scale=0.5]{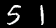} \end{tabular} & 5.99666 & 6\tabularnewline
\begin{tabular}[c]{@{}l@{}} \includegraphics[scale=0.5]{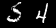} \end{tabular} & 8.6814  & 9\tabularnewline
\bottomrule
\end{tabular}
\end{table}

\begin{table}
\caption{Example results of correct predictions for test set permutations that have the same label but consist of different pairs of digits such as (6, 6) and (4, 8) or (9, 2) and (3, 8). Note also that the model was also able to deal with digits in swapped order, as is the case for (1, 3) and (3, 1). Table~\ref{table:exampleResultsSameLabel2} shows a further example of this.}
\label{table:exampleResultsSameLabel}
\centering
\begin{tabular}[]{@{}lcc@{}}
\toprule
Input Image & Prediction & Actual\tabularnewline
\midrule
\begin{tabular}[c]{@{}l@{}} \includegraphics[scale=0.5]{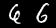} \end{tabular}  & 11.8673 & 12\tabularnewline
\begin{tabular}[c]{@{}l@{}} \includegraphics[scale=0.5]{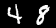} \end{tabular}  & 11.8703 & 12\tabularnewline
\begin{tabular}[c]{@{}l@{}} \includegraphics[scale=0.5]{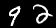} \end{tabular}  & 10.8862 & 11\tabularnewline
\begin{tabular}[c]{@{}l@{}} \includegraphics[scale=0.5]{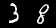} \end{tabular}  & 10.8827 & 11\tabularnewline
\begin{tabular}[c]{@{}l@{}} \includegraphics[scale=0.5]{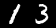} \end{tabular}   & 3.7308  & 4\tabularnewline
\begin{tabular}[c]{@{}l@{}} \includegraphics[scale=0.5]{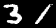} \end{tabular}   & 4.23593 & 4\tabularnewline
\bottomrule
\end{tabular}
\end{table}

\begin{table}[h!]
\caption{Example of correct predictions for three permutations from the same test set that sum to the same value.}
\label{table:exampleResultsSameLabel2}
\centering
\begin{tabular}[]{@{}lcc@{}}
\toprule
Input Image & Prediction & Actual\tabularnewline
\midrule
\begin{tabular}[c]{@{}l@{}} \includegraphics[scale=0.5]{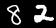} \end{tabular}  & 9.84883 & 10\tabularnewline
\begin{tabular}[c]{@{}l@{}} \includegraphics[scale=0.5]{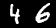} \end{tabular}  & 9.9731  & 10\tabularnewline
\begin{tabular}[c]{@{}l@{}} \includegraphics[scale=0.5]{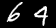} \end{tabular}  & 10.0746 & 10\tabularnewline
\bottomrule
\end{tabular}
\end{table}

Although the network was trained as a regression problem, accuracy can also be measured by rounding the predicted real number output to the nearest integer and comparing it to the integer label. When rounding to the nearest digit, accuracy was as high as 92\%, depending on the train/test split and averaged 70.9\%. Accuracy increases, if the predicted value is used with a floor or ceiling function, and both values compared to the true value, achieving an accuracy of approximately 88\% across a 10 fold cross validation. When allowing for an error of $\pm1$, the accuracy, of course, increases further. Table~\ref{table:crossValAccuracy} shows the accuracy of the trained network over a 10 fold cross validation. The accuracies are measured across all samples of all test set permutations---a total of 10,000 images. The accuracies are provided here merely as a guide, for regression problems it is of course more useful to observe the average loss of the predictions versus the labels. Errors presented here are likely the result of misclassifications of the images themselves rather than the logic of the operator learned, as the network was trained to optimise the loss and not the accuracy. Also, even if overall accuracy would be low, for all permutation pairs there are always correct predictions, again a further reason why it is not entirely useful to report accuracies. The mean squared error loss is provided as a truer measure of the network's performance, and in order to provide as accurate a loss as possible a 10-fold cross validation was performed. The results of a 10-fold cross validation of the permutation pairs can be seen in Table~\ref{table:crossValResults}. The average mean squared error loss over the 10-fold cross validation was 0.853322.

\begin{table}[]
\centering
\caption{Accuracy of each run of a 10-fold cross validation. The accuracies presented here are on a sample-by-sample basis. For each permutation there are always correct predictions, even for poorly performing folds, such as folds 1 and 2. By using floor and ceiling functions on the predicted values for each of the images, the accuracy increases significantly. Note that for the accuracies shown in this table 2,000 samples per permutation were generated.}
\label{table:crossValAccuracy}
\begin{tabular}{@{}llll@{}}
\toprule
   Fold    & Rounding & Floor/ceiling            & $\pm 1$            \\ \midrule
   1       & 81.19\%  & 85.51\%                  & 94.93\%            \\
   2       & 43.00\%  & 90.89\%                  & 96.23\%            \\
   3       & 80.95\%  & 86.95\%                  & 95.12\%            \\
   4       & 55.08\%  & 71.21\%                  & 86.56\%            \\
   5       & 82.35\%  & 93.56\%                  & 95.28\%            \\
   6       & 92.23\%  & 95.49\%                  & 96.76\%            \\
   7       & 54.32\%  & 86.94\%                  & 95.38\%            \\
   8       & 74.86\%  & 94.71\%                  & 96.67\%            \\
   9       & 87.59\%  & 94.33\%                  & 95.91\%            \\
   10      & 57.48\%  & 85.53\%                  & 96.23\%            \\ \midrule
   Avg.    & 70.91\%  & 88.51\%                  & 94.90\%            \\ \bottomrule
\end{tabular}
\end{table}

\begin{table}[]
\centering
\caption{Results of each run of a 10-fold cross validation. The average mean squared error (MSE) across 10 runs was $\approx 0.85$ on the test set.}
\label{table:crossValResults}
\begin{tabular}{@{}lll@{}}
\toprule
 Fold    & Test Set MSE      & Train Set MSE          \\ \midrule
 1       & 1.1072            & 0.0632                 \\
 2       & 0.6936            & 0.0623                 \\
 3       & 0.7734            & 0.0661                 \\
 4       & 0.7845            & 0.0607                 \\
 5       & 0.9561            & 0.0694                 \\
 6       & 0.7732            & 0.0553                 \\
 7       & 1.2150            & 0.0803                 \\
 8       & 0.7278            & 0.0674                 \\
 9       & 0.9464            & 0.0602                 \\
 10      & 0.5556            & 0.0709                 \\ \midrule
 Avg.    & 0.8533            & 0.0656                 \\ \bottomrule
\end{tabular}
\end{table}

\section{Conclusion}
In this work, we have presented a neural network that achieves good results at the task of addition when trained with images of side-by-side digits labelled with their summations, and tested with digit combination pairs it has never seen. The network was able to predict the summation with an average mean squared error of $0.85$ for permutation pairs it was not trained with. By testing the network on a distinct set of digit combinations that were unseen during training, it suggests the network learned the task of addition, rather than a mapping of individual images to labels.
A number of further experiments would be feasible using a similar experimental setup. Most obviously, the use of three digits per image could be performed using permutations up to length 3, or higher. Furthermore, other arithmetic operations could also be tested, such as subtraction or multiplication. More generally, the applicability of this method to other datasets in other domains needs to be investigated more thoroughly, for example whether there is an analogous experiment which could be performed on a dataset that does not involve arithmetic but involves the combination and interpretation of unseen combinations of objects in order to make a classification, such as through the use of the Fashion-MNIST or ImageNet datasets \cite{xiao2017}.

\bibliographystyle{splncs04}
\bibliography{mnist-arithmetic}

\begin{thebibliography}{1}
\providecommand{\url}[1]{\texttt{#1}}
\providecommand{\urlprefix}{URL }
\providecommand{\doi}[1]{https://doi.org/#1}

\bibitem{tensorflow2016}
Abadi, M., Agarwal, A., Barham, P., Brevdo, E., Chen, Z., Citro, C., Corrado,
  G.S., Davis, A., Dean, J., Devin, M., et~al.: Tensorflow: Large-scale machine
  learning on heterogeneous distributed systems. arXiv:1603.04467  (2016)

\bibitem{hoshen2016}
Hoshen, Y., Peleg, S.: Visual learning of arithmetic operations. In: AAAI. pp.
  3733--3739 (2016)

\bibitem{HubelWiesel:1962:CatVisual}
Hubel, D.H., Wiesel, T.N.: Receptive fields, binocular interaction and
  functional architecture in the cat's visual cortex. The Journal of Physiology
   \textbf{160}(1),  106--154 (1962)

\bibitem{lecun1998}
LeCun, Y., Bottou, L., Bengio, Y., Haffner, P.: Gradient-based learning applied
  to document recognition. Proceedings of the IEEE  \textbf{86}(11),
  2278--2324 (1998)

\bibitem{LeCunEtAl:1989:Handwritten}
LeCun, Y., Jackel, L.D., Boser, B., Denker, J.S., Graf, H.P., Guyon, I.,
  Henderson, D., Howard, R.E., Hubbard, W.: Handwritten digit recognition:
  applications of neural network chips and automatic learning. {IEEE}
  Communications Magazine  \textbf{27}(11),  41--46 (1989)

\bibitem{liang2016}
Liang, Z., Li, Q., Liao, S.: Character-level convolutional networks for
  arithmetic operator character recognition. In: 2016 International Conference
  on Educational Innovation through Technology (EITT). pp. 208--212 (2016)

\bibitem{WalachWolf:2016:counting}
Walach, E., Wolf, L.: {Learning to Count with CNN Boosting}. In: Leibe, B.,
  Matas, J., Sebe, N., Welling, M. (eds.) ECCV. pp. 660--676 (2016)

\bibitem{xiao2017}
Xiao, H., Rasul, K., Vollgraf, R.: {Fashion-MNIST: a Novel Image Dataset for
  Benchmarking Machine Learning Algorithms} (2017)

\bibitem{zeiler2012}
Zeiler, M.D.: {ADADELTA:} an adaptive learning rate method. arXiv
  \textbf{abs/1212.5701} (2012)

\end{thebibliography}

\end{document}